\documentclass[OTHER]{cta-author}
\pdfoutput=1 
\usepackage{hyperref}
\usepackage{booktabs}
\usepackage{float}

\newcommand{\grevys}{Gr\'{e}vy's }
\begin{document}

\supertitle{}

\title{Adapting the re-ID challenge for static sensors}

\author{
	\au{Avirath Sundaresan$^{\text{\hspace{2pt}}1,\dag}$}
        \au{Jason Parham$^{\text{\hspace{2pt}}2,\dag}$} 
	\au{Jonathan Crall$^{\text{\hspace{3pt}}2}$} 
        \au{Rosemary Warungu$^{\text{\hspace{2pt}}5}$} 
        \au{Timothy Muthami$^{\text{\hspace{3pt}}5}$} 
        \au{Jackson Miliko$^{\text{\hspace{1pt}}5}$} 
        \au{Margaret Mwangi$^{\text{\hspace{3pt}}5}$} 
        \au{Jason Holmberg$^{\text{\hspace{2pt}}4}$} 
        \au{Tanya Berger-Wolf$^{\text{\hspace{3pt}}6,4}$} 
        \au{Daniel Rubenstein$^{\text{\hspace{2pt}}7}$} 
        \au{Charles Stewart$^{\text{\hspace{2pt}}3,4}$} 
	\au{Sara Beery$^{\text{\hspace{2pt}}8,*}$} 
}

\address{
	\add{1}{Computing and Mathematical Sciences, California Institute of Technology, Pasadena, CA, USA}
	\add{2}{Kitware, Clifton Park, NY, USA}
	\add{3}{Rensselaer Polytechnic Institute, Troy, NY, USA}
        \add{4}{Wild Me, Conservation X Labs, Portland, OR, USA}
        \add{5}{Laikipia Zebra Project, Mpala Research Centre, Laikipia, Kenya}
        \add{6}{Computer Science and Engineering, Ohio State University, Columbus, OH, USA}
        \add{7}{Ecology and Evolutionary Biology, Princeton University, Princeton, NJ, USA}
        \add{8}{Electrical Engineering and Computer Science, Massachusetts Institute of Technology, Cambridge, MA, USA}
        \add{\dag}{Co-authors}
	    \email{beery@mit.edu}
}



\begin{abstract}
The \grevys zebra, an endangered species native to Kenya and southern Ethiopia, has been the target of sustained conservation efforts in recent years. Accurately monitoring \grevys zebra populations is essential for ecologists to evaluate ongoing conservation initiatives. Recently, in both 2016 and 2018, a full census of the \grevys zebra population has been enabled by the Great \grevys Rally (GGR), a citizen science event that combines teams of volunteers to capture data with computer vision algorithms that help experts estimate the number of individuals in the population. A complementary, scalable, cost-effective, and long-term \grevys population monitoring approach involves deploying a network of camera traps, which we have done at the Mpala Research Centre in Laikipia County, Kenya. In both scenarios, a substantial majority of the images of zebras are not usable for individual identification, due to "in-the-wild" imaging conditions --- occlusions from vegetation or other animals, oblique views, low image quality, and animals that appear in the far background and are thus too small to identify. Camera trap images, without an intelligent human photographer to select the framing and focus on the animals of interest, are of even poorer quality, with high rates of occlusion and high spatio-temporal similarity within image bursts. We employ an image filtering pipeline incorporating animal detection, species identification, viewpoint estimation, quality evaluation, and temporal subsampling to compensate for these factors and obtain individual crops from camera trap and GGR images of suitable quality for re-ID. We then employ the Local Clusterings and their Alternatives (LCA) algorithm, a hybrid computer vision {\&} graph clustering method for animal re-ID, on the resulting high-quality crops. Our method processed images taken during GGR-16 and GGR-18 in Meru County, Kenya, into 4,142 highly-comparable annotations, requiring only 120 contrastive same-vs-different-individual decisions from a human reviewer to produce a population estimate of 349 individuals (within 4.6$\%$ of the ground-truth count in Meru County). Our method also efficiently processed 8.9M unlabeled camera trap images from 70 camera traps at Mpala over two years into 685 encounters of 173 unique individuals, requiring only 331 contrastive decisions from a human reviewer.

\end{abstract}
\maketitle



\section{Introduction}
\label{sec:intro}

The population of \grevys zebras experienced a dramatic decline beginning in the 1970s largely due to hunting and competition for food and water resources with local pastoral communities. Estimates have placed the number of \grevys zebras remaining in the wild at under 2,000, with the vast majority in the Samburu region of central Kenya. Due to extensive conservation efforts by the Kenyan and Ethiopian governments as well as environmental NGOs, the population of \grevys zebras have stabilized in recent years \cite{rubenstein_equus_2016}.

Accurately censusing \grevys zebra populations is critical for ecologists to evaluate these existing conservation efforts. Developing and maintaining a census of known individuals requires effective animal re-identification methods to ensure that only unique individuals are included in population counts \cite{Tuia2022-hm}. A popular method for population size estimation is ``capture-mark-recapture" \cite{Pradel1996-xn}. By this method, a set of animals in the target population is first captured and marked, then a second set of animals is independently recaptured, and finally a population estimate formed from the number of animals captured twice. This method, however, proves difficult to scale to large populations and territories, and may lead to inaccurate population estimates when animals are not confined to the study area and
\makebox[\linewidth][s]{are capable of evading tagging \cite{Lee2014-pu,Tilling1999-ri}. Further, the tagging methods} \par

\begin{figure}[H]
\vspace{-3pt}
\centering{\includegraphics[scale = 0.24]{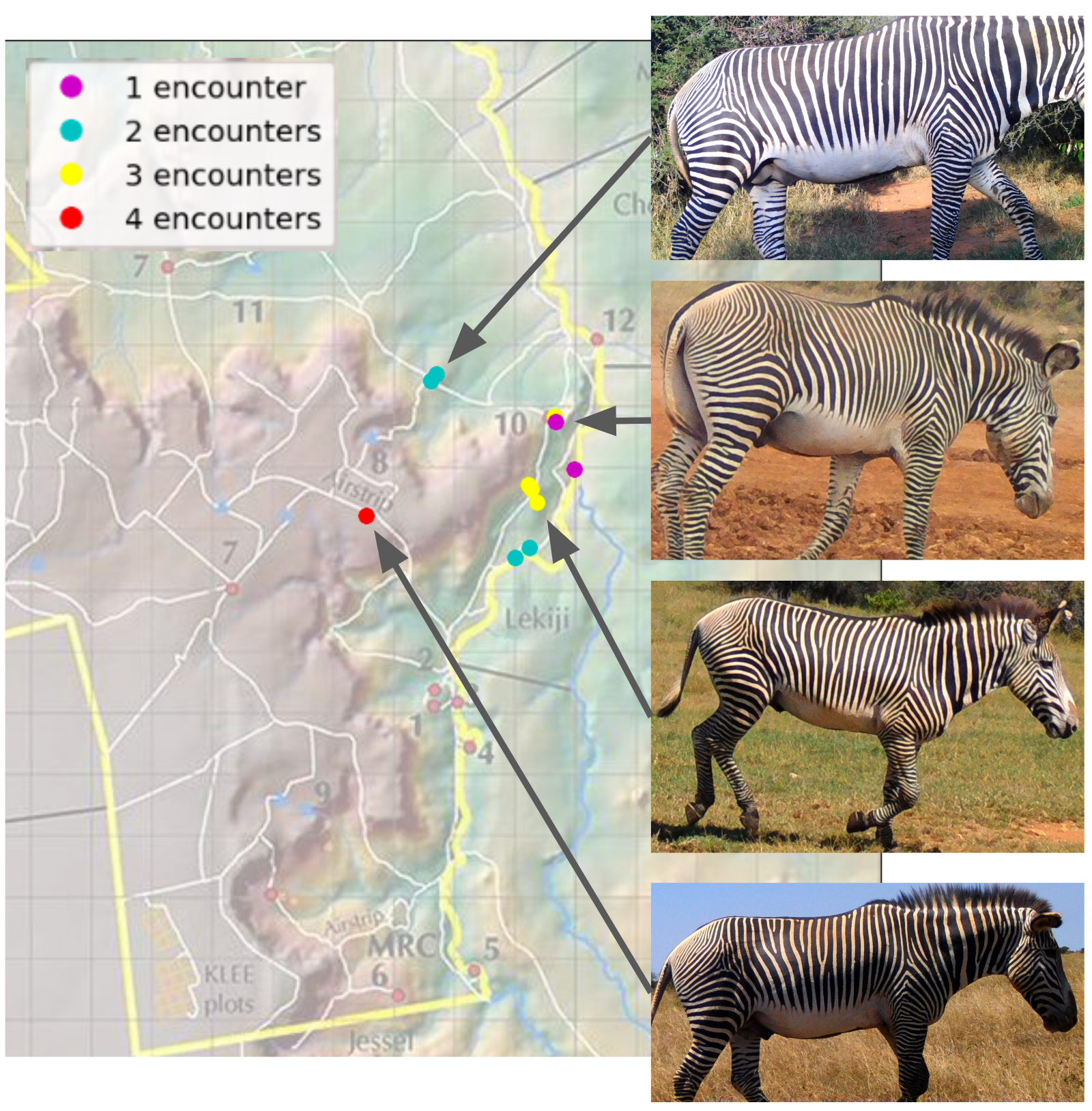}}
\caption{Map of several encounters of a single individual within Mpala Research Centre identified from camera trap data by our method and requiring minimal human labeling. \label{fig:ind32}}
\end{figure} 

\noindent used in population estimation studies (such as ear tags and radio collars \cite{Alexander1994-wz,Thouless1995-dq,Collins2014-mt}) can be excessively expensive and time-consuming to implement in the field, and overly harmful to the animal \cite{Witmer2005-hg}. 

A modern alternative to manual mark-recapture studies that does not have these limitations is exemplified in the Great \grevys Rally (GGR) events of 2016 and 2018 where volunteers spread over the range of \grevys zebra to photograph as many animals as possible over two consecutive days. Relying on the distinctive appearance of an animal itself as a means for identification, a combination of algorithmic and human curation efforts produced population estimates that have been accepted as definitive by the Kenyan Wildlife Service and the IUCN \cite{rubenstein_state_2018}. Despite the success of the GGR, the curation effort of \char`\~50,000 images per event is formidable, since the dataset must be narrowed to the recognizable subset of images of \grevys and a streamlined curation procedure must be developed, all-the-while preserving the accuracy of the resulting census.  ~\cite{Parham_Diss} demonstrates how to ensure accuracy while reducing human effort.

\begin{figure*}[!t]
\centering{\includegraphics[scale = 0.28]{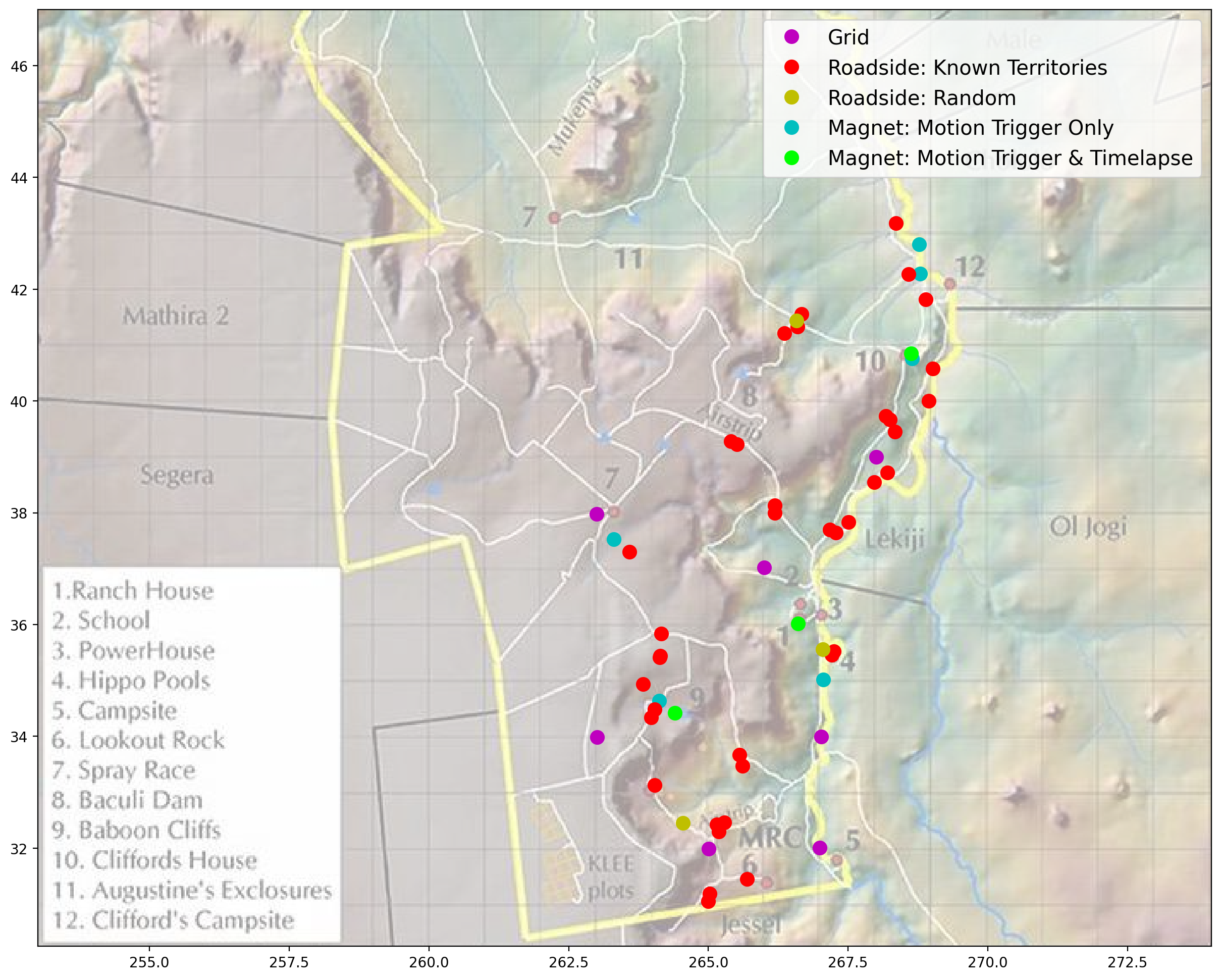}}
\centering{\includegraphics[scale = 0.28]{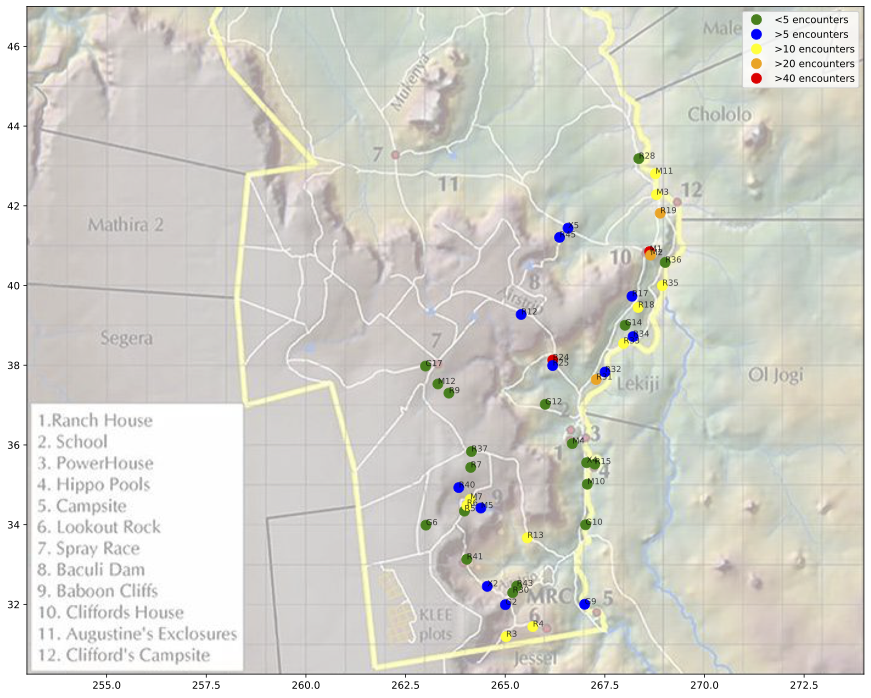}}
\caption{\textbf{(Left)} Location of our camera traps across the Mpala Research Centre. These cameras are placed according to four distinct strategies: random grid, randomly along roads, at known \grevys territories along roads, and at magnets such as salt licks and watering holes. \textbf{(Right)} Map of all \grevys encounters across Mpala. \label{fig:enc_locs} \label{fig:cam_locs}}
\vspace{-10pt}
\end{figure*}

Complementary to the focused burst of human effort for the GGR photographic events is the possibility of using a fleet of camera traps to collect the data needed for population monitoring. For animals that allow for sight-based ID, images taken by camera traps are a potentially cost-effective and non-invasive method to re-identify individuals and obtain robust population estimates. Unlike alternative methods such as tagging or even the field photography of the GGR events, camera traps do not require the physical presence of field researchers for animal re-identification; this ensures that the natural behaviors of animals are not disrupted and saves considerable time and resources \cite{Delisle2021-vt,Schneider2019-wz}.

However, analysis of camera trap images by human observation alone is not practical. In recent years, computer vision algorithms have been shown to present a highly accurate and standardized method for camera trap image analysis. Computer vision techniques have seen significant success in automated species detection {\&} identification from camera trap images, especially with advances in deep learning \cite{Norouzzadeh2019-fo,Beery2019-rc,Beery2019-qt,Beery2019-ma,Tabak2020-ti}. Accurate species ID from camera trap images has laid the foundation for the next step in an end-to-end photographic censusing pipeline from camera trap images: automated individual re-ID (for reviews, see: \cite{Schneider2019-wz,Ravoor2020-ff,Vidal2021-cg}). However, the challenging nature of the automatically-captured data (motion blur, occlusion, poor lighting, far-away animals) often leads to flawed animal re-identification even by human experts \cite{meek2013reliability}. Additionally, images captured in a burst from the same motion trigger are often highly repetitive, leading to potential bias in automatic re-identification if not handled carefully.

In this work, we seek to adapt existing techniques to the combined challenges of animal re-identification in GGR-style rally events and in the especially challenging static camera trap paradigm.

Unlike many other classification tasks with numerous and highly similar classes, animal re-identification is an open-set classification problem \cite{geng2020recent}, with the need to assign every unique individual to its own class and to recognize novel individuals unseen during training. There are two key categories of animal re-ID algorithms. \textit{Ranking} algorithms for re-ID query an image of the target individual against an existing database to obtain a ranking of the most confident matches. Hotspotter \cite{Crall13} is one such texture-based ranking algorithm that uses the SIFT \cite{Lowe2004-wi} algorithm to extract salient features from the query image and subsequently a nearest neighbor search to match the query image against the database. The algorithm is specialized for striped and spotted animals, and it has been used for re-ID of \grevys zebra \cite{Crall13,Rubenstein_GGR} and several other animal species from camera trap data \cite{Park2019-vh,Nipko2020-kb,blount_comparison_2022}. \textit{Verification} algorithms for re-ID, on the other hand, do not require querying an existing database; instead, a verification algorithm simply decides whether two images contain the same individual. An example of a verification algorithm is the Verification Algorithm for Match Probabilities (VAMP) \cite{Crall_Diss}, a random forest classifier \cite{Breiman2001-py,Pal2005-ww} that receives two images and decides if they contain the same animal, different animals, or are incomparable. Lastly, contrastive deep learning algorithms, such as the Pose Invariant Embedding (PIE) network \cite{Zheng2017-yp,Moskvyak2021-vd}, can learn a global feature embedding (instead of handcrafted features with Hotspotter \& VAMP) for a particular image, allowing for distance-based comparisons with the feature embeddings of other images in a database; this allows PIE to serve as both a ranking and verification algorithm simultaneously. However, unlike Hotspotter and VAMP, PIE (and deep learning algorithms in general) requires a significant amount of training data. Instead, when considering a population without individual-level ground truth labels readily available, it is preferable to use classical computer vision for ranking and verification that can be easily bootstrapped with minimal human supervision. 

Equally important to the choice of re-ID algorithms is determining which annotations --- bounding boxes within images drawn around \grevys zebras by a detection algorithm --- should even be considered for re-ID.  Annotations that are of poor quality, are partially occluded, are from an indistinguishable viewpoint, or show an uncommon viewpoint lead to significantly increased work by humans and can cause over-counting. To avoid this, we introduce the notion of a "census annotation" to restrict the attention of re-ID algorithms to annotations that should be universally recognizable.  

In this paper, we use census annotations, Hotspotter and VAMP in conjunction with the Local Clusterings and their Alternatives (LCA) decision management algorithm \cite{stewart_LCA_in_prep, Parham_Diss} for \grevys zebra re-ID. Like other animal and human re-ID systems \cite{Kulits2021-of,Bodesheim2022-zu,Wang2016-lj}, the LCA algorithm offers a human-in-the-loop approach to dynamically cluster annotations by individual and request human reviews for verification hard cases. This paper is an extended version of work presented at the 3rd International Workshop on Camera Traps, AI, and Ecology \cite{Sundaresan2023}. We extend our previous work by broadening the scope to include human-captured images from the Great Grévy’s Rally alongside static camera trap data and elaborate on our methodology for ensuring annotation comparability across encounters. To the best of our knowledge, this marks the first use of interactive {\&} error-driven clustering algorithms like LCA for animal re-ID for either events like the GGR or camera trap data for any species. 

\section{Methods}
\label{sec:methods}

\subsection{GZCD Dataset}

\begin{figure}[!t]
\centering{\includegraphics[scale = 0.18]{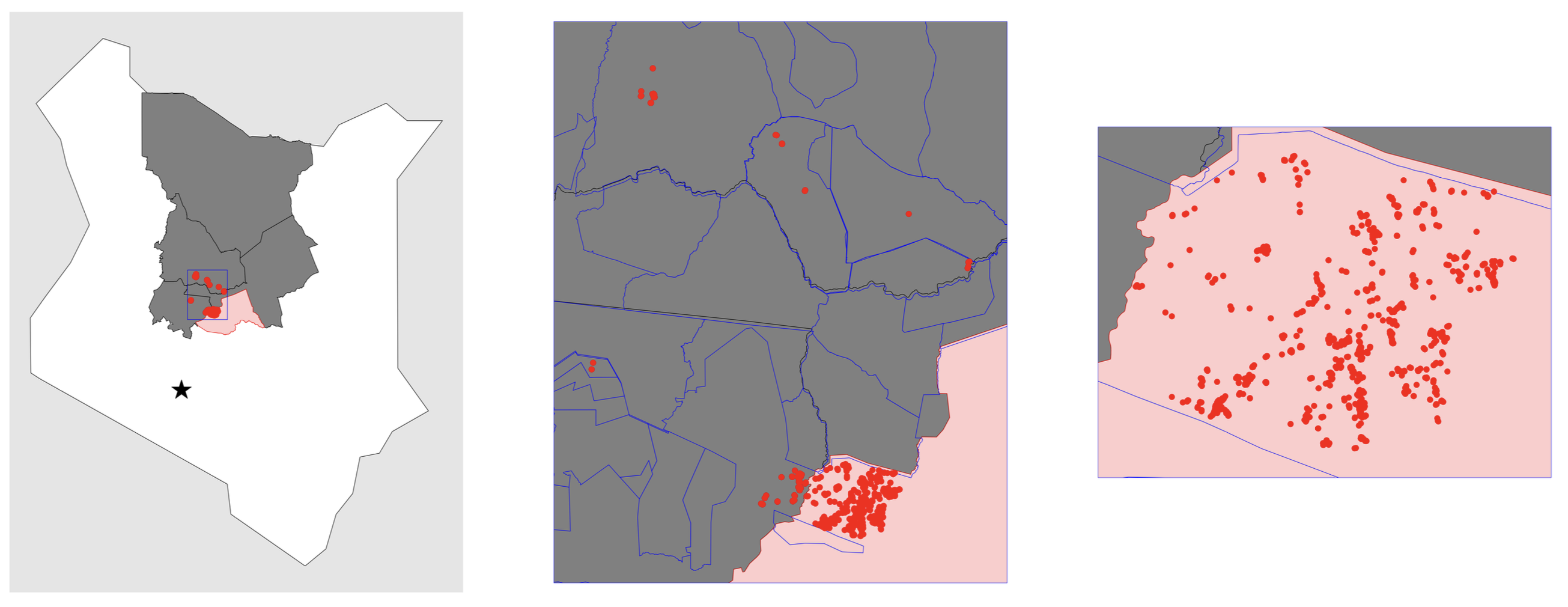}}
\caption{Map of image GPS locations in the GZCD dataset. Meru County, Kenya (in red) is located north of the capital (star) and is at the base of Mt. Kenya. Includes all images by photographers that took images in Meru County, even if they were not taken in that county.\label{fig:gzcd}}
\vspace{-21pt}
\end{figure}

The GZCD dataset is comprised of 5,464 images sourced from Meru County, Kenya, taken over four days of GGR-16 and GGR-18 by 13 photographers (see Fig. \ref{fig:gzcd}). The photographers were trained to capture a consistent right-side viewpoint for \grevys zebra, and only images taken of the intended side were kept in the dataset by human reviewers. The dataset is highly curated: bounding boxes and labels (species, viewpoint, and quality) were set by human reviewers for all animals. The 7,372 right-view \grevys annotations were further filtered based on these quality labels, resulting in a set of 4,269 ``quality baseline" right-view \grevys annotations. Note that the spatial subset for Meru County, Kenya is geographically isolated by mountains from neighboring conservation areas, giving the expectation that the population is largely self-contained. Thus, the GZCD dataset is an excellent testbed for our censusing pipeline.

\subsection{Camera Trap Dataset}

We use images collected from a network of 70 camera traps distributed around the Mpala Research Centre in Kenya’s Laikipia Plateau. The network has collected 8.9 million images over the past two years of deployment. There are four types of camera trap placement schemes in the network: systematically in a grid, at ``magnet" sites (e.g. salt licks), as well as expert-targeted \& random placement along roads (see Fig. \ref{fig:cam_locs}).

\begin{figure*}[!t]
\centering{\includegraphics[scale = 0.3]{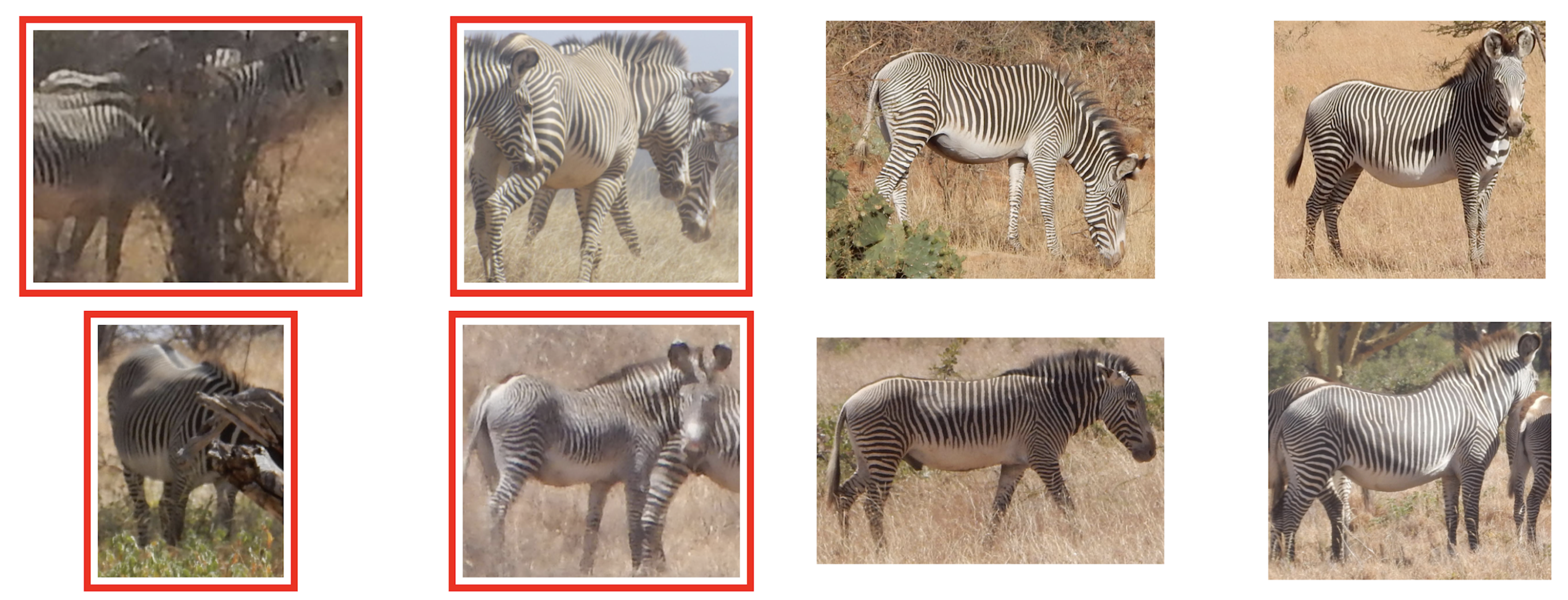}}
\caption{Example images of Census Annotations for Grévy’s zebra. Annotations on the left (boxed in red) were marked as non-CAs by a reviewer, and annotations on the right were marked as CAs.  \label{fig:ca_example}}
\vspace{-10pt}
\end{figure*}

\subsection{Automated Species Identification}

Prior to re-ID, the raw camera trap images were first passed through a YOLO v2 species detection model \cite{Parham2018-pw} to localize all zebra with a bounding box (both \grevys and plains zebra species), and crop the region of the image within each bounding box for downstream use. Localized bounding boxes crop out irrelevant and potentially distracting background information and yield distinct \textit{annotations} of independent individuals from images that feature several animals. YOLO v2 has shown to be more accurate for animal detection than alternative object detection models, such as Faster R-CNN \cite{Parham2016-nd}. Next, the cropped regions are classified to zebra species --- \grevys vs.\ plains zebra --- and viewpoint --- left vs.\ right --- by a DenseNet model. Only right, front-right, and back-right viewpoints are considered for identification; differing viewpoints cannot be matched with one another, as the right and left sides of Grevy’s zebra are distinct. Both the annotation localization and classification networks were trained on the WILD dataset \cite{Parham2018-pw}.

\subsection{Census Annotations}

Beyond assigning viewpoint and species to each annotation, we wish to ensure that these annotations — from both human-captured and camera trap images — would be \textit{universally comparable}.  In principle, when such an annotation fails to match the other annotations we should be able to safely conclude that it is the only annotation from that animal. Furthermore, focusing on these annotations (a) should make verification decisions easier both for an algorithm such as VAMP and for a human reviewer ("verifier"), (b) should allow recovery from mistakes such as incidental matching \footnote{As examples, a matching algorithm may recognize animals seen in the background (i.e., a ``photobomb''~\cite{Crall_Diss}); camera traps are acutely affected by a similar scenario by accidentally matching background textures}, and (c) should avoid increased human effort to address matching ambiguities.\footnote{See time and effort studies of animal ID curation in ~\cite{Parham_Diss}}. For \grevys zebras, this entails annotations that have both the distinctive hip and shoulder chevron (see Fig. \ref{fig:ca_example}, which we will call ``census annotations" (CA) \cite{Parham_Diss}. 

\begin{figure}[!t]
\centering{\includegraphics[scale = 0.25]{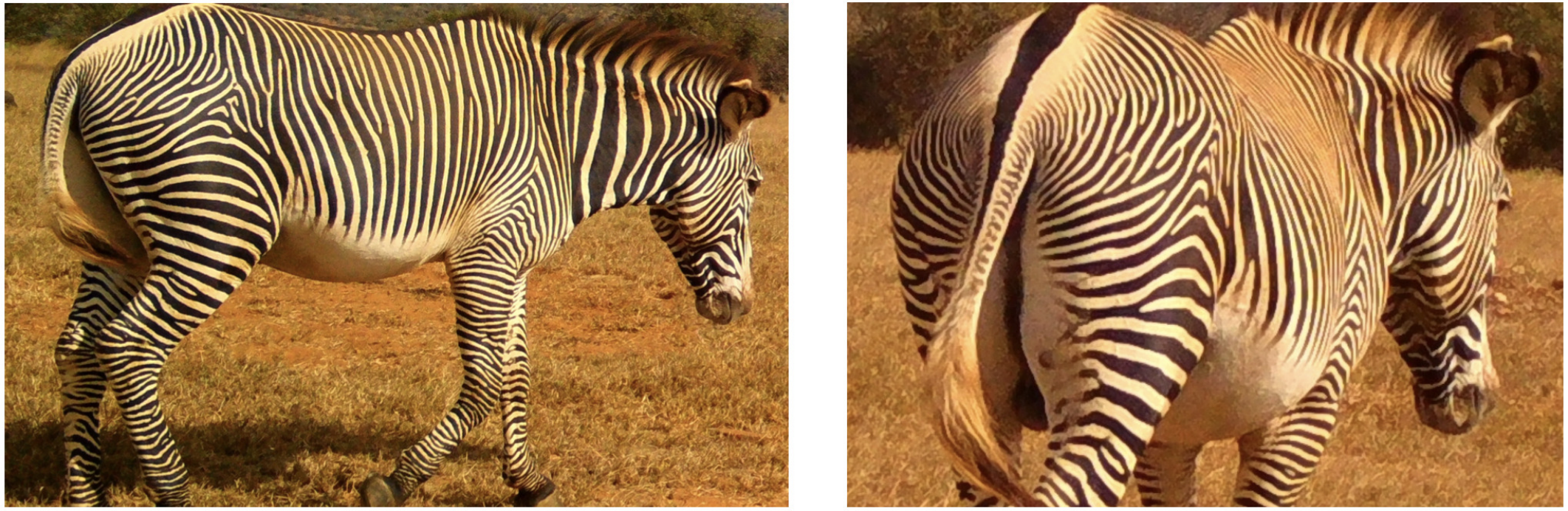}}
\caption{Example of a good and poor quality CA of an individual within an encounter. \textbf{(Left)} Annotation with CA score of 0.9997; the hip and shoulder chevrons are very clearly visible. \textbf{(Right)} Annotation with CA score of 0.0032; viewpoint is correct but the hip and shoulder chevrons are not clearly visible.\label{fig:ca_comp}}
\end{figure}

In order to train a census annotation network, we started by curating 10,229 \grevys zebra annotations from GGR-18. These were arranged in a series of grids and presented via a web interface to two reviewers, who selected the annotations that were \grevys CAs. Then, a DenseNet with a linear classification layer was trained on this CA dataset to decide whether an annotation was indeed a CA, producing an associated CA confidence score (see Fig. \ref{fig:ca_comp}). Finally, a regression network was trained to narrow the annotation region to only surround the hip and shoulder chevron and minimize any distracting background information. Each of these new annotations was saved as a ``Census Annotation Region'', or CA-R \cite{Parham_Diss}. The CA and CA-R networks were then used to obtain the CA-Rs and corresponding CA scores for both GZCD and the camera trap dataset. 

\subsection{Filtering Pipeline}

\begin{figure*}[!t]
\centering{\includegraphics[width=.9\linewidth]{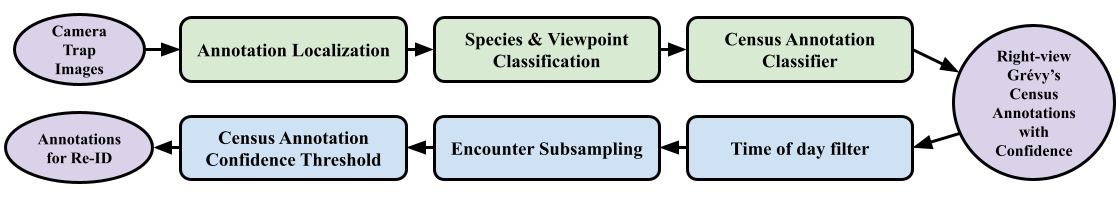}}
\caption{Data curation pipeline from unlabeled camera trap images to high-quality individually identifiable census identifications. Data is represented in purple, models in green, and quality filtering processes in blue. The annotations remaining after this curation pipeline next have individual features extracted using VAMP which are used as input into the LCA algorithm. \label{fig:pipeline}}
\end{figure*}

Mark-recapture statistical models are highly sensitive to inaccurate identifications, but are built with the assumption that not all individuals are seen. Thus, it is important to prioritize the precision of our re-identification pipeline over assigning an ID to each annotation. One significant factor in inaccurate identification, by both humans and algorithms, is image quality --- blurry, or poorly lit annotations are difficult or impossible to reliably identify. Filtering annotations by quality is therefore a critical component of the re-ID pipeline, and automated quality filtering reduces the amount of human time and work required to rectify matching errors caused by undesirable annotations. Here, we devise an annotation filtering scheme suitable for re-ID from camera traps (Fig. \ref{fig:pipeline}). 

As already described, we only consider census annotation regions (CA-Rs), those the show right-side viewpoints of \grevys zebra, including both the shoulder and hip chevron. Beyond this, we further filter by time of day: only annotations from images taken between 6:30 AM and 7:00 PM (sunrise and sunset at Mpala) were kept for re-ID. Due to the camera trap settings, images taken during the daytime are optical (RGB) and are of higher resolution (13 MP). Nighttime images are taken with an infrared flash at lower resolution (9 MP). Qualitatively, nighttime images are more difficult for human reviewers than daytime images; without modification, we believed the VAMP verification algorithm was likely to perform better on higher contrast and quality images taken during the day. 

\begin{figure}[!t]
\centering{\includegraphics[width=.9\linewidth]{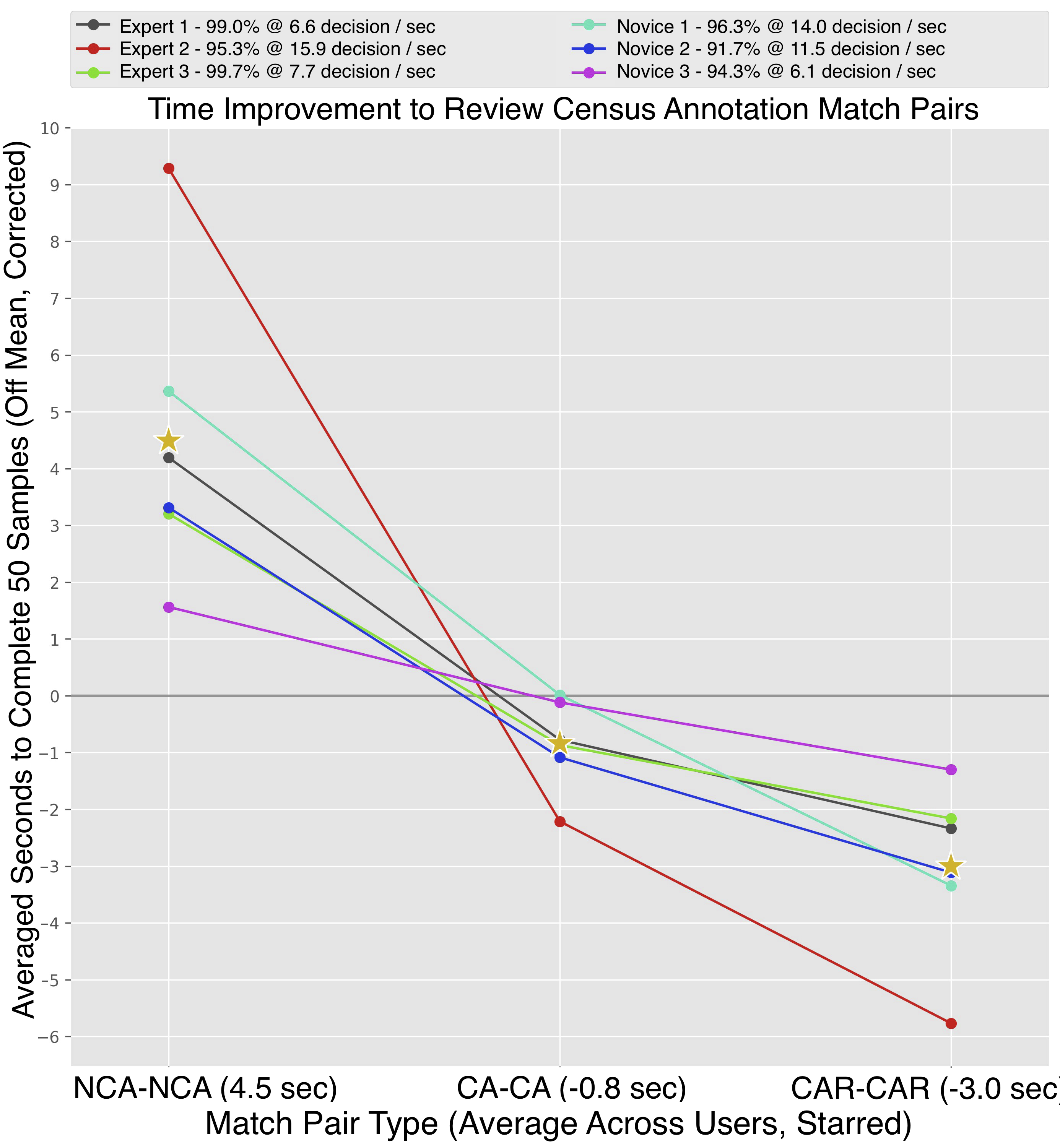}}
\caption{A user study from~\cite{Parham_Diss} demonstrating that users are faster and more accurate at reviewing decisions with CA and CA-R.  Beyond the savings in human effort by automating more of the review, the overall effort is reduced further by making decisions easier for novice and expert reviewers. \label{fig:user-study}}
\end{figure}

Next, the annotations were filtered by \textit{encounter}. To define the encounters, we used an agglomerative clustering approach: for each camera, annotations from images taken within the same minute and in consecutive minutes were grouped together in the same encounter. Next, the annotation with the highest Census Annotation confidence score was selected from each encounter, and the rest were discarded. As the camera traps take images in bursts, the images in a encounter are nearly identical, and hence the corresponding annotations very likely must feature the same individuals. This step encourages matching across the best representatives from every encounter, reducing the amount of data required but preserving the relevant information. Experiments in \cite{Parham_Diss} show that filtering in this manner, in particular filtering for CA-Rs, does not have a significant impact on the individual count, but dramatically reduces the human effort required to produce it (Fig. \ref{fig:user-study}). While the GZCD used mobile photographers, the encounters it used were based on location and time, and often included many Census Annotations; the camera trap dataset used a shorter time-only decision to reduce trivial duplicates.

Lastly, the annotations were filtered for CA confidence score. Annotations above a 0.31 confidence score threshold were kept for re-ID. In previous work \cite{Parham_Diss}, a lower threshold was shown to add incomparable annotations to the re-ID database and encourage spurious matches, while a higher threshold was shown to eliminate relevant sightings without a considerable increase in matching speed and accuracy. The filtering pipeline yielded 685 right-view \grevys zebra annotations across the reserve, with each representing a single encounter (see Fig. \ref{fig:enc_locs}). 

\subsection{Zebra ID Curation}

The goal of the LCA decision management algorithm is to build an \textit{identification graph} with the filtered annotations, using automated ranking \& verification and delaying human intervention as much as possible. In the identification graph, annotations are represented as vertices, and relationships between annotations (whether positive or negative matches) are represented as edges. The LCA algorithm seeks to group annotations into clusters corresponding to individuals by maximizing positive edge weights within clusters and negative weights between clusters. This simplifies the identity labeling process significantly: instead of being asked to identify an annotation as one of up to 150 known individuals, reviewers are asked to compare pairs of annotations and determine whether they belong to the same individual or not. This contrastive task can be accomplished even by non-experts due to innate human capacity for pattern matching --- it is similar to a game of spot the difference. 

The ID graph is initialized with no edges, with every annotation constituting its own cluster. For every annotation, the Hotspotter ranking algorithm returns a list of its most confident matched annotations, and LCA forms edges between the annotation and its potential matches. Next, the VAMP verification algorithm evaluates every pair of vertices connected by an edge and assigns an edge weight (positive or negative) based on its confidence in the match. The LCA algorithm has two main phases. In the \textit{scoring} phase, it keeps the initial edges and weights intact. LCA iterates through every local clustering (a single cluster or a pair of clusters) and checks to see if there exists an alternative clustering that increases a score measure defined as the sum of intra-cluster edge weights minus the sum of inter-cluster edge weights. If such an alternative clustering exists, the new arrangement is adopted. Once there are no longer any better local alternative clusterings, LCA proceeds to the \textit{stability} phase. In this phase, LCA considers local clusterings for which the difference between it and its next best alternative are sufficiently small that changes in edge weights can potentially introduce a superior alternative clustering. For such local clusterings, LCA requests additional reviews to VAMP and the human (if VAMP cannot litigate the review on its own) to determine if certain edge weights must change. Once all local clusterings are significantly better than their next best alternative, the algorithm is considered to have converged. 

\section{Results}

\subsection{Results on GZCD}

\label{sec:results-GZCD}

As discussed above, there were 4,269 ``Quality Baseline'' right-view \grevys annotations selected for identification from GZCD. Additionally, from the initial set of 7,372 right-view \grevys annotations in the GZCD, 4,142 were classified as CAs by our CA model using a score quality threshold of 0.31. Furthermore, each of these CAs were manually annotated to yield 4,142 corresponding CA-Rs. It should be noted that the CAs are not a strict subset with the Quality Baseline set; the two sets overlap significantly for trivially good sightings but each have annotations that are not present in the other. These three annotations sets — quality baseline, CAs, and CA-Rs — were each passed into the LCA algorithm for identification. 

As the GZCD is composed of images taken during GGR-16 and GGR-18 in Meru County, we are able to compare the ID curation results with ground-truth estimates obtained from the two rallies.  The Lincoln-Petersen index~\cite{petersen_yearly_1896} for the ``Quality Baseline'' is $360\pm27$ in 2016 and $399\pm29$ for 2018. The population estimate based on only CA was $366\pm27$ for GGR 2016 and $373\pm29$ for GGR 2018, whereas the CA-R estimate was $366\pm27$ zebra in 2016 and $373\pm29$ animals in 2018~\cite{parham_great_2018}. 

With the Quality Baseline set, the LCA algorithm requested 22,972 decisions in total, with 420 from the human reviewer (yielding an automation rate of 98.2\%). However, LCA converged faster with the CA set, with 18,607 total number of reviews, 352 of which were performed by the human reviewer (yielding an automation rate of 98.1\%). We observe that LCA converged with 4.4 decisions per annotation with the CA set compared to 5.3 with the quality baseline set, indicating that the census annotations were more discriminative and required fewer alternative clusterings for LCA to consider. Finally, LCA converged the fastest with the CA-R set, with 13,427 reviews requested in total and only 120 performed by the human reviewer (yielding an automation rate of 99.1\%). The results with the CA-R set were accurate within 4.6\% on GGR-16 data (predicting 349$\pm$26 animals against a ground-truth value of 366$\pm$27) and within 0.5\% (371$\pm$30 animals predicted against 373$\pm$29 ground-truth). In comparison, the quality baseline LCA results underestimated the ground-truth counts by only 12 individuals for GGR-16 and 1 individual for GGR-18, but required nearly four times the number of human reviews -- therefore, considering CA-Rs relative to the quality baseline results in a 71.4\% reduction in human work, while the resulting estimates still remain within the confidence interval of the established baseline count. These results are summarized in Table \ref{tab1}. While the CA-R result was obtained with a major reduction in human effort, Figure~\ref{fig:user-study} also demonstrates that the amount of time spent on each CA-R review is substantially decreased (and more accurate).

\subsection{Results on Camera Trap Dataset}
\label{sec:results-CT}
8.9 million camera trap images from the initial dataset were passed through the species ID pipeline, yielding 84,383 zebra images (including both plains and \grevys zebra). Following species and viewpoint filtering, 23,512 right-view \grevys zebra annotations remained across 3,338 distinct encounters. After excluding nighttime encounters, 1,138 daytime encounters remained. We then sampled the single highest-quality annotation from each of the 1,138 encounters, as determined by our CA model, to avoid matching on near-identical images from the same burst. We further used a score quality threshold of 0.31 on census annotations that reduced the number of annotations to be identified to 734. As a last step, we used a blurriness filter to remove any additional challenging annotations, resulting in a final set of 685 CA-Rs. 

\begin{figure}[h!]
\centering{\includegraphics[width=\linewidth]{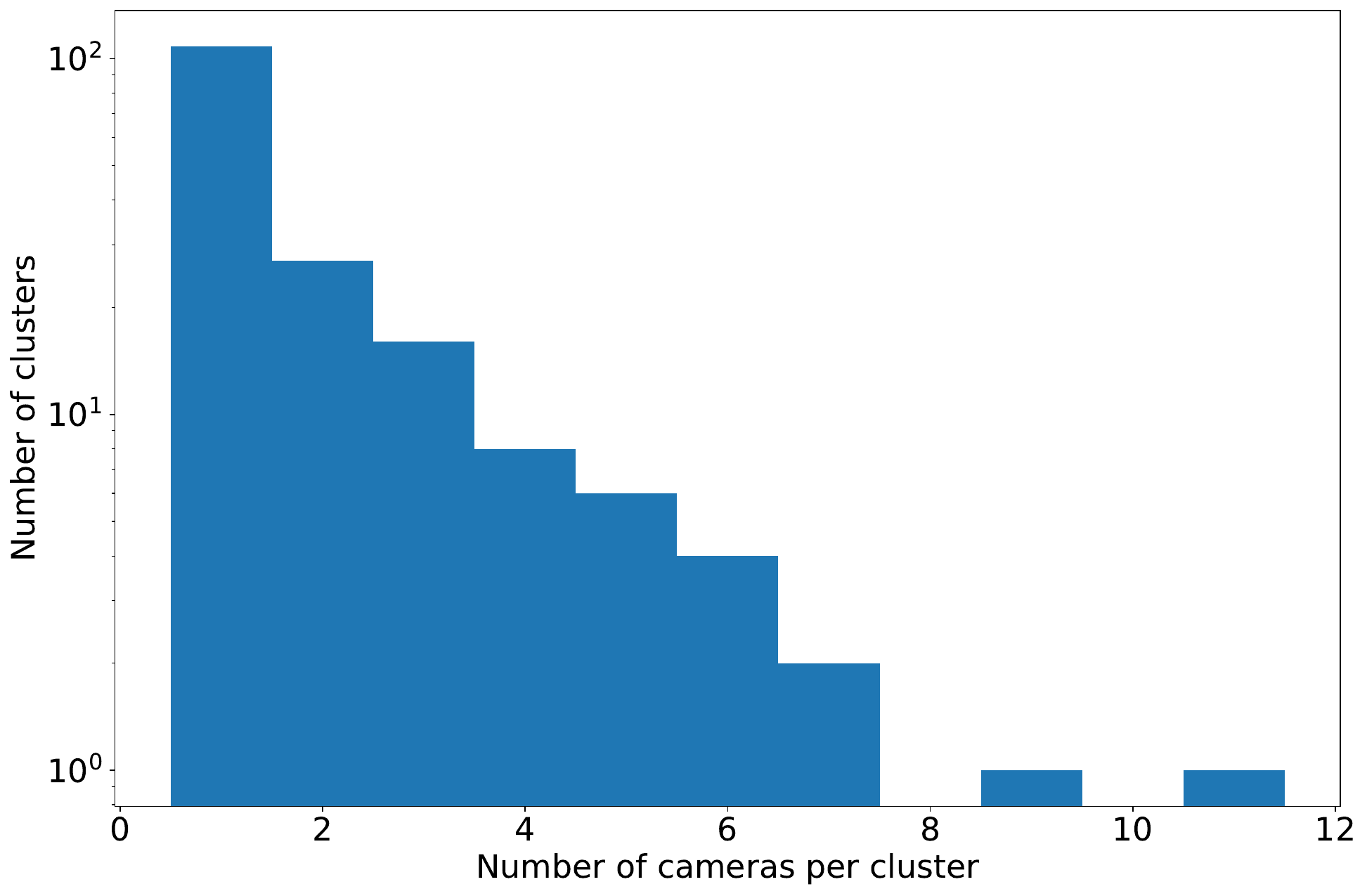}}
\caption{Distribution per individual of distinct camera locations where the individual was seen, emphasizing that our method is able to overcome the static background bias and identify individuals in different contexts. \label{fig:cams_per_cluster}}
\end{figure}

\begin{figure}[H]
\centering{\includegraphics[width=\linewidth]{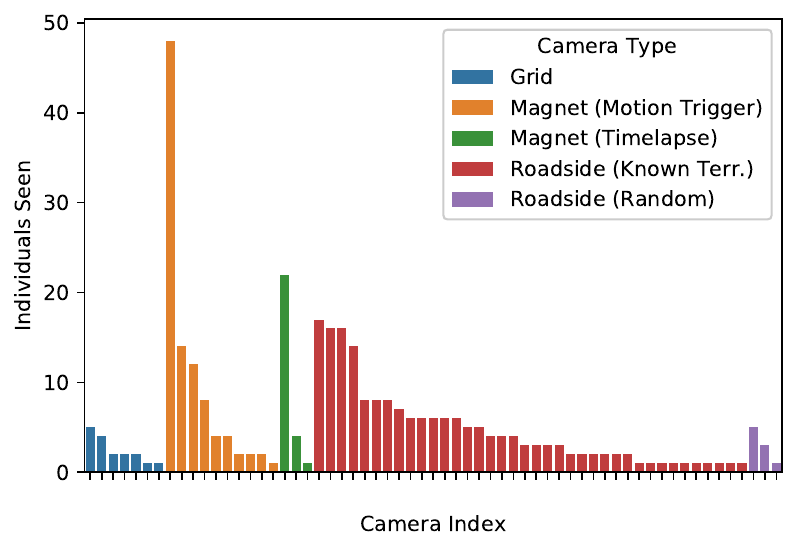}}
\caption{Distribution of individuals sighted per camera, sorted by placement strategy. A distinction is made here between motion-trigger and timelapse cameras at magnet sites. 
\label{fig:indivs_per_strat}}
\end{figure}

In order to associate these 685 annotations to clusters of individuals, the LCA algorithm requested 5,403 automated reviews by VAMP and 331 additional human reviews; pairwise comparisons of annotations were performed largely automatically using VAMP, with a automation rate of 93.9\%. The converged ID graph had 173 clusters, each corresponding to an individual zebra ID within the 685 annotations. On average, each individual was sighted across 1.9 static camera traps and 3.96 encounters, indicating significant matching across space and time. See Supplementary Fig. \ref{fig:encs_per_cluster} for the distribution of encounters for each individual, and Supplementary Fig. \ref{fig:cams_per_cluster} for the distribution of static camera traps that sighted each individual. The number of individuals sighted by each camera also differed based on its placement strategy (random grid, known \grevys territories along roads, randomly along roads, and both timelapse \& motion-trigger traps at magnet sites). On average, camera traps at a magnet site (e.g. salt lick, dam, etc.) sighted the most individuals on average (8.5 per trap) and also produced the most \textit{first} individual sightings on average (5.85 per trap). See Table \ref{tab2} and Supplementary Fig. \ref{fig:indivs_per_strat} for the distribution of total individuals sighted by cameras of each placement strategy. Particularly, the encounters for large individual clusters were spread across many months, with some spanning more than a year (see Supplementary Fig. \ref{fig:time_dist}). Based on the distinct encounters and cameras associated with each individual, we can generate spatial maps of encounters for any individual across the reserve: see Fig. \ref{fig:ind32} for a map of encounters for a particular zebra (Individual 32).

\section{Discussion}
\label{sec:discuss}

Census Annotations and Census Annotation Regions are critical to automated photographic census because they 1) speed up human verification of match pairs and reduce the number of manual decision errors, 2) better separate the positive and negative scores predicted by algorithmic verifiers like VAMP, 3) reduce the amount of incidental matching~\cite{Parham_Diss}, and 4) drastically reduce the amount of human interaction needed during manual review. Census Annotation Regions are powerful because they force a photographic census to consider only the most critical information for matching.  Furthermore, our results indicate that using Census Annotations and Census Annotation Regions results in consistent population estimates on both the known GZCD baseline and camera trap dataset.  

For our static camera dataset, the proposed combined method of subsampling and curation has enabled us to efficiently combine the 685 high-quality encounters to form 173 clusters, each representing an individual \grevys zebra. The use of LCA for semi-automatic, interactive decision-making resulted in fewer than 0.5 human reviews per annotation. By contrast, when working only with ranking algorithms, typically several --- five to ten --- potential matches must be examined, and LCA includes consistency checking implicitly. 

\begin{table}[t]
\processtable{LCA review requests for GZCD \label{tab1}} 
{\begin{tabular*}{20pc}{@{\extracolsep{\fill}}llllll@{}}\toprule
    \begin{tabular}[c]{@{}l@{}} Set \end{tabular} & 
    \begin{tabular}[c]{@{}l@{}}Annotations\end{tabular} & 
    \begin{tabular}[c]{@{}l@{}}VAMP\\ reviews\end{tabular} & 
    \begin{tabular}[c]{@{}l@{}}Human\\ reviews\end{tabular} & 
    \begin{tabular}[c]{@{}l@{}}Total\\ reviews\end{tabular} & 
    \begin{tabular}[c]{@{}l@{}}Automation\\ rate\end{tabular} \\ 
    \midrule
Quality Baseline & 4,269 & 22,552 & 420 & 22,972 & 98.2\% \\
CA & 4,142 & 18,255 & 352 & 18,607 & 98.1\% \\
CA-R & 4,142 & 13,307 & 120 & 13,427 & 99.1\% \\ 
\botrule
\end{tabular*}}{}
\end{table}

\begin{table}[t]
\processtable{Number of individuals sighted by camera placement strategy \label{tab2}}
{\begin{tabular*}{20pc}{@{\extracolsep{\fill}}llll@{}}\toprule
    & Total & Avg. & New \\ \midrule
Random grid & 16  & 2.29 & 1.43 \\
Roadside, known territories & 105 & 2.76 & 2.26 \\
Roadside, random & 9 & 3 & 0.33 \\
Magnet & 107 & 8.5 & 5.85 \\ \botrule
\end{tabular*}}{}
\end{table}

Each annotation in the cluster is representative of an encounter with that individual at a known position and time, and enables us to convert a set of 8.9 million unlabeled camera trap images into identified encounters that can be used as the input into a spatially-explicit mark recapture model \cite{green2020spatially} to estimate the total \grevys population. Because camera traps are monitoring long-term, in this case over two years unlike the contained 2-day period captured by each Great Grevy's Rally, assumptions about no births and no deaths and no transients in the population are invalid. It is important to note that the ecologists at the Laikipia Zebra Project have confirmed that the \grevys population at Mpala has been an open set over the period in question: of the estimated 150 \grevys individuals currently seen over repeat seasons at Mpala, only 4 are ``stable" and reside on the reserve long-term, while the vast majority (130 expert re-identified individuals) have been seen to migrate and return periodically, driven by water scarcity and threat of predation \cite{KWS}. It is thus likely that the camera network captured some number of transient individuals over the two year sampling span. 

Limitations of the method may include imperfect CA labeling, where annotations with seemingly high CA scores are in actuality difficult to match, leading to two (or more) clusters in place of a single cluster. Additionally, we observe that errors could also occur via potential failures in any one of the algorithmic components whether at the species ID or at the individual re-ID level. Human decision failures in particular may inflate the number of individual clusters, with ambiguous matches mislabeled by the human reviewer resulting in more clusters than necessary. A significant focus of our ongoing work is to further analyze these results in order to tease these factors apart.

Lastly, while the method is currently limited to images taken during daylight hours, an approximately equal-sized set of images and annotations is collected at night. In recent preliminary tests, we selected the twenty-five night time annotations having the best combination of census annotation score and image contrast, and matched them against the clustered daytime annotations produced as a result of the work described here. For twenty of them (80\%) Hotspotter produced correct matches in the top few. This shows potential for extending the overall method to the night-time, but many details must be revisited due to the overall lower quality of nocturnal images before they can be reliably included in generating enhanced census results.

\section{Conclusion}
\label{sec:conclusion}
In this paper, we perform efficient semi-automated \grevys zebra re-ID from camera trap data, with an algorithmic tool chain using Hotspotter for ranking, VAMP for verification, and LCA for decision management. Our method can be used even by non-zebra experts \cite{Parham_Diss}, as it only requires contrastive comparisons between pairs of individuals instead of matching each individual into the previously identified population. Ultimately, this curational re-ID process found 173 distinct individuals in camera trap data collected across a two year period at the Mpala Research Centre. 

Going forward, we aim to further refine the filtering pipeline to improve matching with fewer human reviews and reduce attrition of data through the pipeline. We also are excited to further explore the results of the method, for example by adjusting the CA score threshold and observing how this affects LCA convergence behavior {\&} results. We also hope to further adapt LCA to enable temporal sequences and spatial relationships to be taken into account, to enable us to better make use of the additional images of the individuals from each camera trap image burst and the spatial structure of the habits of individuals over time. Finally, and along similar lines, we hope to introduce ``short-circuiting" to reduce human reviews when spatio-temporal constraints imply that there is no possibility of a match. 




\bibliographystyle{unsrtnat}
\bibliography{references}
\newpage \clearpage
\section*{Supplementary Figures}
Additional data visualization and analysis figures have been included in this section to give additional context to the reader.

\begin{figure}[h]
\centering{\includegraphics[width=\linewidth]{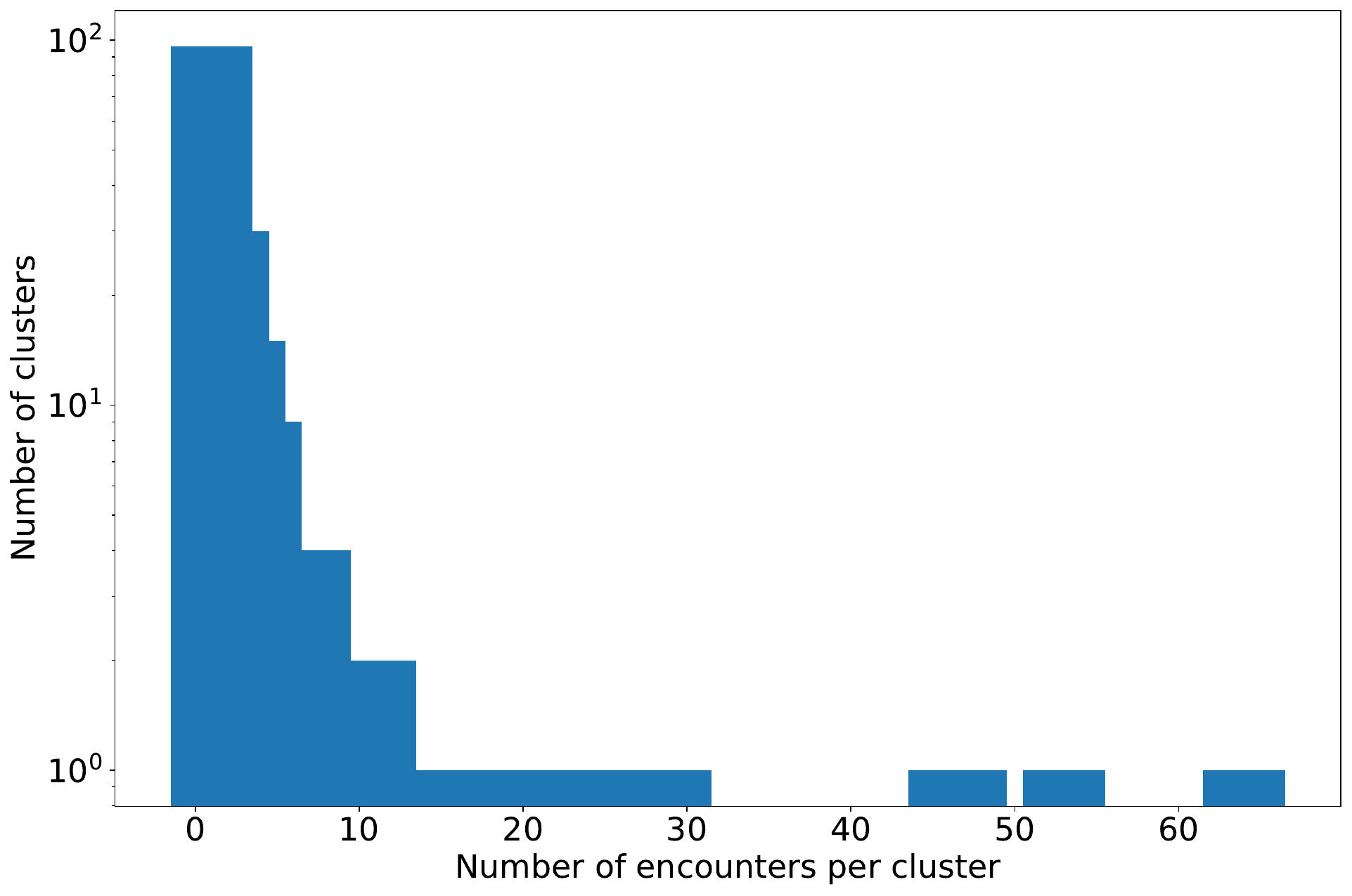}}
\caption{Distribution of number of encounters per individual cluster. \label{fig:encs_per_cluster}}
\end{figure}

\begin{figure}[h!]
\centering{\includegraphics[width=\linewidth]{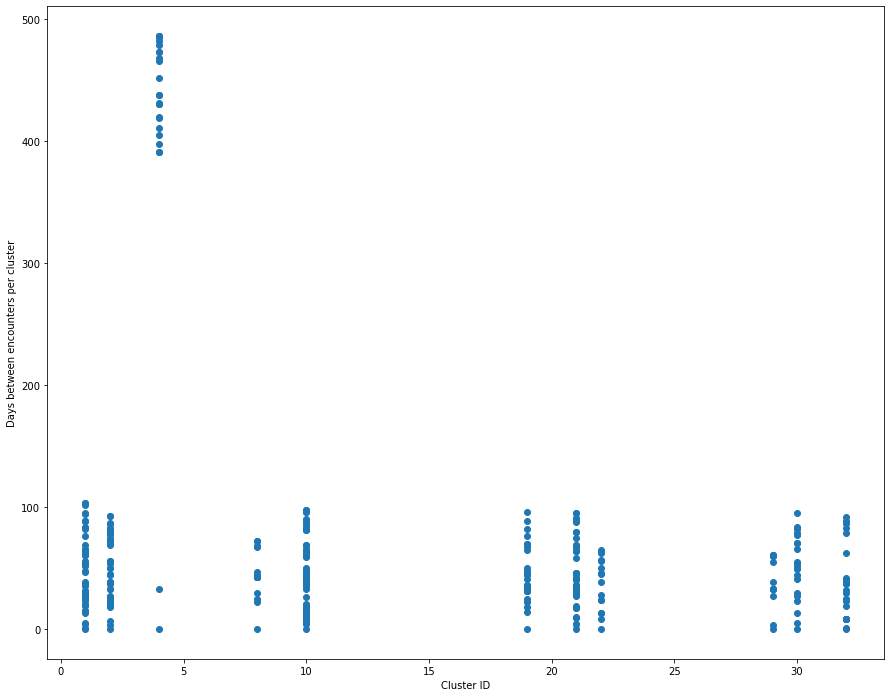}}
\caption{Time in days between encounters for the largest clusters in the ID graph. Note that our method re-identified one of the individuals \textasciitilde 400 days later, after it had migrated out of Mpala and then returned. 
\label{fig:time_dist}}
\end{figure}

\clearpage
\end{document}